# Target-Dependent Multimodal Sentiment Analysis Via Employing Visual-to-Emotional-Caption Translation Network using Visual-Caption Pairs

Ananya Pandey, Dinesh Kumar Vishwakarma[*]
Biometric Research Laboratory, Department of Information Technology, Delhi Technological University, Bawana Road, Delhi-110042, India
Ananyapandey_2k21phdit08@dtu.ac.in, dinesh@dtu.ac.in[*]

**Abstract**
The natural language processing and multimedia field has seen a notable surge in interest in multimodal sentiment recognition. Hence, this study aims to employ Target-Dependent Multimodal Sentiment Analysis (TDMSA) to identify the level of sentiment associated with every target (aspect) stated within a multimodal post consisting of a visual-caption pair. Despite the recent advancements in multimodal sentiment recognition, there has been a lack of explicit incorporation of emotional clues from the visual modality, specifically those pertaining to facial expressions. The challenge at hand is to proficiently obtain visual and emotional clues and subsequently synchronise them with the textual content. In light of this fact, this study presents a novel approach called the Visual-to-Emotional-Caption Translation Network (VECTN) technique. The primary objective of this strategy is to effectively acquire visual sentiment clues by analysing facial expressions. Additionally, it effectively aligns and blends the obtained emotional clues with the target attribute of the caption mode. The experimental findings demonstrate that our methodology is capable of producing ground-breaking outcomes when applied to two publicly accessible multimodal Twitter datasets, namely, Twitter-2015 and Twitter-2017. The experimental results show that the suggested model achieves an accuracy of **81.23%** and a macro-F1 of **80.61%** on the Twitter-15 dataset, while **77.42%** and **75.19%** on the Twitter-17 dataset, respectively. The observed improvement in performance reveals that our model is better than others when it comes to collecting target-level sentiment in multimodal data using the expressions of the face.

## 1   Introduction

In the contemporary era of digital connectivity, an enormous amount of publicly accessible multimodal content is posted by individuals on prominent websites like YouTube, LinkedIn, Twitter, Instagram, Facebook, etc. Analysing such multimodal information streams is advantageous in examining individuals' emotional responses towards a particular topic. Aspect-Based Sentiment Analysis (ABSA) determines the sentiment polarity towards specific attributes that are explicitly mentioned within a given input text. For example, *"The river gives a pleasant view, however, the quality of the roll was disappointing"*. The user conveys a positive opinion regarding the $river's$ location while expressing dissatisfaction or negative sentiment for the target aspect $'roll'$.

\* Corresponding Author
Email: ananyapandey_2k21phdit08@dtu.ac.in (Ananya Pandey),
dinesh@dtu.ac.in (Dinesh Kumar Vishwakarma)

Previous research studies have revealed numerous techniques to perform sentiment recognition tasks for the target aspects. The conventional methods for addressing this challenge primarily involved the creation of comprehensive, hand-crafted features, which were subsequently fed into a classifier. The field of aspect-based sentiment classification has seen notable progress due to the widespread adoption of deep learning techniques in the domain of NLP. Various deep neural network designs have been put forward to address this task, such as [1], [2], [3], [4] and [5]. In recent years, numerous studies have aimed to enhance the representation of semantic interactions between target aspects and the context words by integrating attention modules in conjunction with several recurrent neural networks (RNNs) [6], [7], [8], [9], and [10].

With the growing popularity of multimodal information across social media, it has become inadequate to rely just on written text for the purpose of aspect-based sentiment categorisation. Multimodal postings often include visual elements, such as photos and emoji, which may frequently provide a significant understanding of people's emotions. The opinion towards a target aspect, expressed by a user, is often influenced by the accompanying image. This is because the textual information in such posts is typically unstructured, informal and brief. Consider **Figure 1** as an example. When omitting the accompanying image, the anticipated sentiments towards the targets "Justin," "America," and "Lydia" seem to be neutral, positive, and neutral, respectively. However, this prediction is inaccurate. In the instances mentioned above (**Figure 1**), users convey their sentiments towards "Justin," "America," and "Lydia" by utilising distinct visual representations. Specifically, a crying face is employed to express negative sentiment towards Justin, a neutral image is used to convey a neutral feeling towards America, and a happy look is employed to denote a positive view towards Lydia.

| Visual | Textual | Target | Sentiment |
|---|---|---|---|
| 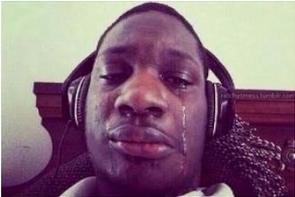 | RT @ irauhlcarlyrae : Justin tweeted # mybeliebers so Beliebers trended # ourJustin | Justin | **Negative** |
| 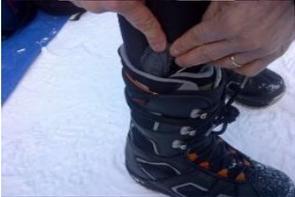 | Randy from America knows how to wear a good sock. Hide it under your boot for extra warmth. # tweetusyoursocks | America | **Neutral** |
| 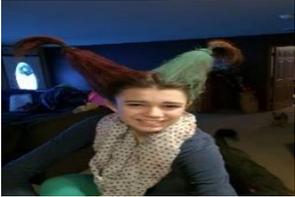 | Crazy hair day! Lydia is a contender. :) | Lydia | **Positive** |

**Figure 1** A few instances of Target-Dependent multimodal sentiment recognition are provided, including the identified targets and their corresponding sentiments.

Previous studies [11], [12], [13],[14],[15], and [16] have proven that the region of the face depicted within visual content serves as a robust indicator of emotion. The study [16] discovered that (1) during the observation of visual cues, human observers exhibit a greater attentional bias towards affective objects than neutral items, and (2) this bias towards emotional objects is of greater significance when objects are associated with humans. According to statistical information, it has been observed that more than 60% of the visual content in any corpus derived from Twitter displays facial emotions. Hence, we suggest that considering facial expressions is crucial in Target-Dependent Multimodal Sentiment Analysis (TDMSA). Hence, we believe the explicit use of facial information as visual indicators of emotions to convert visual, emotional cues into textual form facilitates cross-modal fusion.

Motivated by the recent advancements [17], [18], [19], and [20] made in understanding facial emotion within the domain of computer vision, we provide an effective and simple approach, i.e., visual-to-emotional-text translation network for aspect-based multimodal sentiment recognition. This technique aims to convert the sentiment conveyed in the image into a textual representation by generating descriptions of facial expressions. As outlined in **Figure 2**, the primary contributions of our study can be briefly summed up as follows:

1. We have developed a novel framework called the Visual-to-Emotional-Caption Translation Network (VECTN) to perform Target-Dependent Multimodal Sentiment Analysis (TDMSA). This network is composed of three main modules: the facial emotion description module, the target alignment and refinement module for face description, and the fusion module.
2. The facial emotion description unit is responsible for generating a face description that includes various features such as age, gender, and emotion. The target alignment and refinement module estimates the cosine similarity between the visual input and the face descriptions with the target. In the fusion component, two robustly optimised pre-trained language models are utilised to simulate images, captions and face descriptions by a gating mechanism for feature fusion and noise reduction.
3. We perform extensive studies and experiments utilising standard datasets, namely Twitter-2015 and Twitter-2017, to demonstrate the effectiveness and reliability of our approach in simulating multimodal representations. Through our research, we aim to achieve remarkable cutting-edge results.

## 2 Related work
### 2.1 Target-Dependent Unimodal Sentiment Recognition

The classification of sentiment at the aspect level, sometimes referred to as target-dependent sentiment recognition, is a crucial task of sentiment recognition that has received significant research interest in recent years [21,22]-[23]. The vast majority of existing methodologies may be broadly classified into two main categories.

One area of study involves utilising techniques such as Part-of-Speech Tagger and lexicons of the sentiments to create a customised feature. These features are then used in conjunction with conventional statistical learning methods to make predictions about sentiment. Although these models have shown reasonable performance on various standard datasets [24], [25], [26], and [27], their effectiveness is limited by their extensive dependence on feature engineering.
Another area of research focuses on the integration of target information into different deep-learning-based architectures. *Ruder et al.* [1] introduced a Bi-LSTM-based hierarchical architecture to leverage inter- and intra-sentential dependencies. Later, *Peng et al.* [2] proposed a two-stage approach that combined aspect extraction, sentiment recognition, and explanation of the reason for the sentiment into a single task. In this approach, LSTM is employed to predict the emotional state of the target phrases after retrieving the aspect terms. In addition to deep learning models, attention-based frameworks have gained significant interest among academics as a means to address the challenges associated with text-based sentiment analysis. *Phan et al.* [6], *Li et al.* [10], and *Hoang et al.* [7] utilised different variants of BERT, a large language attention-based model, for the prediction of sentiment at the target level. *Liu et al.* [8] designed two innovative attention modules, namely, context and content attention, to deal with the issue of semantic discrepancy. The concept of context-based attention is utilised to consider both the sequential arrangement of words and their interdependencies. On the other hand, content-based attention has the ability to retrieve pertinent information for a specific target by taking into account the information included within the full phrase, thereby providing an overall view. Yang *et al.* [9] integrated a dedicated co-attention module with the layers of Bi-LSTM. This approach enables the network to acquire non-linear representations of both the target and context in a more efficient manner, leading to improved sentiment prediction. Motivated by the benefits of attention strategies in achieving large contextual information, many research efforts have proposed diverse attention mechanisms [28] and [29] to effectively represent the relationships between the context and its target.

However, most of these research efforts have primarily concentrated on illustrating the textual context by assessing its significance to the target aspect. Yet, they have failed to account for visual characteristics that have become increasingly prevalent in the era of social media.

### 2.2 Target-Dependent Multimodal Sentiment Recognition

In recent years, with the rise of multimodal content in online communities, data derived from multiple modalities (audio, visual, etc.) has been employed to fuse with conventional textual attributes to predict the sentiment more accurately. Research in the domain of multimodal sentiment recognition [30] involves the utilisation of visual-captions pairs to conduct target-dependent analysis.

The concept of aspect-based multimodal sentiment analysis was initially developed by *Yu et al.* [31]. The two publically accessible multimodal benchmark datasets released in this research study comprise pairs of visual-caption, specifically referred to as "Twitter-2015" and "Twitter-2017". The study conducted by *Yu et al.* [31] presented the TomBERT framework, which consists of a ResNet branch for extracting visual features from photographs, two BERT branches for extracting textual features from captions and targets, and a fusion mechanism to integrate all the feature representations for the purpose of sentiment prediction. Subsequently, *Xu et al.* [32] introduced the first Chinese multimodal dataset, "Multi-Zol," designed explicitly for entity-based sentiment recognition. This research paper developed the MIMN network, which combines Bi-GRU and ConvNet and integrates them with attention modules for extracting textual and visual features. The ESAFN architecture, proposed by *Yu et al.* [33], is a composite model that combines LSTM, ResNet-152, and attention techniques to forecast sentiment at the entity level. The multi-headed attention-based framework and the ResNet-152 architecture are adopted by *Gu et al.* [34] to address textual and visual data. The objective of integrating the capsule network and multi-headed attention is to capture the relationship between multiple modalities effectively. *Zhang et al.* [35] also utilised a similar approach, namely the integration of LSTM with ResNet, for target-dependent sentiment analysis by fusion of different attentional networks. To effectively capture the most significant visual information from the dataset, *Zhao et al.* [36] developed an intra and intermodal relationship graph ConvNet. These graphs will facilitate the analysis of the caption and visual data. In *Wang et al.* [37], after the visual-caption pairs have been processed using Bi-GRU and a ConvNet, the resulting fused vector of both representations is passed to the BERT module. This module is responsible for modelling the effective interaction between the inputs. *Yu et al.* [38] introduced a hierarchical transformer model to effectively capture the interaction between caption-target and visual-target. Recently, *Xue et al.* [39] proposed a MAMN framework. This framework comprises three stages: the first stage focuses on feature extraction from multiple modalities, the second stage aims to enhance the representation of multimodal information by reducing noise, and the final stage addresses the fusion strategy. [40], [41], [42], and [43] are a few other research publications in the field of Target-Dependent Multimodal Sentiment Recognition.

Despite the latest advancements in target-dependent sentiment recognition, the comprehensive understanding of information cannot be accurately predicted solely based on textual data. Existing approaches primarily leverage object-level semantic details in images while neglecting the visual modality's explicit utilisation of facial emotional clues. One of the primary challenges that has to be addressed is the distillation of emotional clues retrieved from the visual modality and effectively integrating them with the textual information. Hence, this paper presents a Visual-to-Emotional-Caption Translation Network (VECTN) method that concentrates on acquiring visual-emotional facial clues and aligning and combining them with the target entity in textual modality.

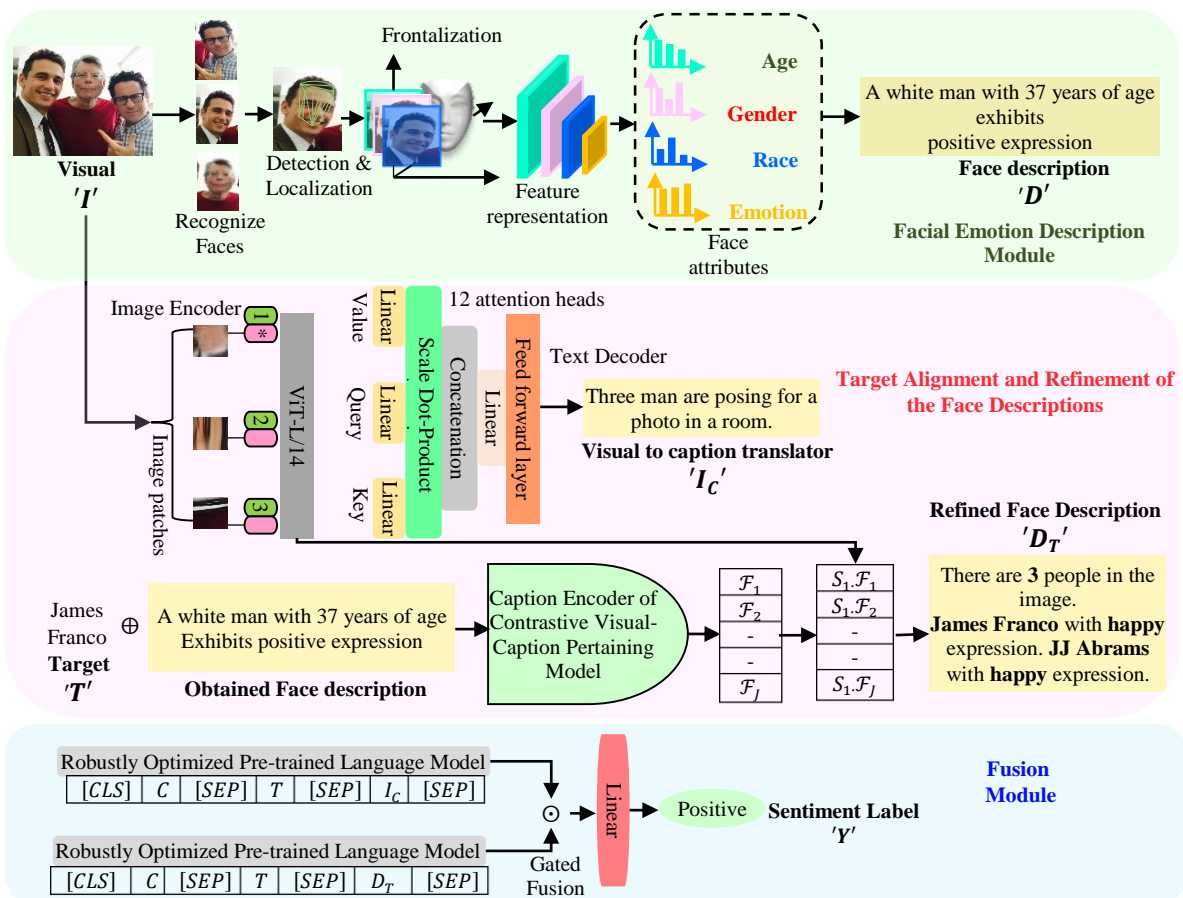

**Figure 2** Proposed Methodology ('⊕' denotes concatenation and '⊙' denotes gated fusion).

**Table 1** Listed below are some important symbols and their respective meanings.

| Symbols | Meaning |
|---|---|
| $E$ | Multimodal example |
| $I$ | Visual modality available in the dataset |
| $C$ | Caption modality available in the dataset |

| Symbols | Meaning |
|---|---|
| $T$ | Target entity available in the dataset |
| $Y$ | Output sentiment labels |
| $\mathcal{F}$ | Total number of faces extracted from the visual modality |
| $D$ | The set represents the face description obtained from $\mathcal{F}$. |
| $D_T$ | Refined face descriptions concatenated with the target entity |
| $I_C$ | Set of captions generated by using visual modality $I$ of the dataset |
| $O_{D\&T}$ and $O_I$ | Embedding of visual and textual content |
| $V$ and $b$ | Trainable parameters (Weights & Bias) |
| $\oplus$ | Denotes concatenation |
| $\odot$ | Denotes gated fusion |

## 3 Proposed Methodology

### 3.1 Problem Formulation

The TDMSA can be precisely described as outlined below: Consider a collection of visual-caption pair examples denoted as $M = \{E_1, E_2, E_3, \ldots\ldots, E_M\}$, where $|M|$ represents the total number of instances. For each given example, an image $I \in R^{3 \times H \times W}$ is provided where 3, $\mathcal{H}$ and $\mathcal{W}$ indicate the number of channels, height and width. Every visual sample in this study is associated with textual content represented by a set of $K$- words provided by the captions $C = \{w_1, w_2, w_3, \ldots\ldots, w_K\}$, which comprises a subsequence of $N$-word that represents the target entity, defined as $T = \{w_1, w_2, w_3, \ldots\ldots, w_N\}$. Our study aims to develop a sentiment classifier to predict a sentiment label $Y$ from multimodal examples accurately. A combination of variables $E = \{I, C, T\}$ represents each sample in $E$. The sentiment labels are categorised into three classes: $Y \in \{positive, negative, neutral\}$.

### 3.2 Visual-to-Emotional-Caption Translation Network (VECTN)

The proposed framework Visual-to-Emotional-Caption-Translation Network (VECTN) illustrated in **Figure 2** has four distinct components: Facial emotion description module, Target alignment and refinement of the face descriptions, Image captioning, and Fusion module. For a given tweet consisting of a visual-caption pair, denoted as $E = \{I, C, T\}$, we first take the input image $'I'$ and feed it into a facial emotion description unit to generate face description $D = \{D_1, D_2, D_3, \ldots\ldots, D_\mathcal{F}\}$ comprises of different features such as age, gender, emotion, etc., where $\mathcal{F}$ is the number of faces present in an input image and $D_i = \{D_1, D_2, D_3, \ldots\ldots, D_L\}$ represents a phrase consisting of $L$-word. The extraction and textualisation of facial expressions within an image, which represents an immense amount of information on the individual's sentiments, is the primary emphasis of this module. Since the input image $'I'$ may include several facial expressions, it is necessary to match or align the facial description $D_T$ with the target entity $T$. The target alignment and refinement of the face descriptions module estimates cosine similarity between visual input $I$ and face descriptions with target $D_T$. The facial description $D_T$ is selected and rewritten based on similarity scores. Since the visual scenes may provide extra semantic details, we employ the image-to-text transformer (*Wang et al.* [44], 2022) to produce image captions for the scene $I_C = \{I_{C1}, I_{C2}, I_{C1}, \ldots\ldots, I_{CG}\}$, where $G$ represents caption length. At last, in the fusion component, we employ two robustly optimised pre-trained language models based on BERT to simulate image captions and face descriptions, followed by a gating mechanism for feature fusion and noise reduction. For target-dependent sentiment recognition, the gated unit output flows via a linear layer. The pseudocode for the proposed algorithm is presented in **Table 2**. The following subsections will provide detailed insights for every module. **Table 1** presents a comprehensive collection of significant symbols along with their respective meanings, aiming for a better understanding of the proposed strategy.

### 3.2.1 Facial Emotion Description Module

The proposed module addresses two fundamental difficulties in TDMSA. First, the complex images in multimodal tweets might make it challenging to extract object-level emotional indicators. Another issue is translating emotional signals obtained from visual modality into a sequence of words.

To address the first challenge, as previously stated, leveraging the wide range of facial expressions in images proves to be an efficient method for extracting emotional cues from visual mode. The first step involves using a tool represented as $\oint$ developed by *Serengil et al.*[45] to recognise multiple faces within an image of the dataset as stated in **Equation (1)**. Let $\mathcal{F}$ represent the set of faces, denoted as $\mathcal{F} = \{\mathcal{F}_1, \mathcal{F}_2, \mathcal{F}_3, \ldots\ldots\ldots, \mathcal{F}_J\}$, where $J$ represents the total number of faces and $\mathcal{F}_J \in \mathcal{R}^{3 \times \mathcal{H}_\mathcal{F} \times \mathcal{W}_\mathcal{F}}$ represents a face area with *three* channels, $\mathcal{H}_\mathcal{F}$ height, and $\mathcal{W}_\mathcal{F}$ width. The obtained faces $\mathcal{F}$ are then fed into a pre-trained classification model (*Serengil et al.* [46]) for facial attribute analysis, which involves gender, age, race (Indian, Black, Asian, White, Latino, and Middle Eastern,) and facial expression to predict sentiments.

$$\mathcal{F} = \oint (I) \tag{1}$$

**Table 2** pseudocode for the proposed algorithm

| |
|---|
| **Algorithm 1:** Target-Dependent Sentiment Recognition based on visual-to-emotional-caption translation network (VECTN-Net) |
| **Aim:** To learn a mapping function $F: (I, C, T, Y) \rightarrow$ from the multi-modal training examples $E$. |
| **Input:** visual set $I \in \mathcal{R}^{3 \times \mathcal{H} \times \mathcal{W}}$; <br>     Caption set $C = \{w_1, w_2, w_3, \ldots\ldots\ldots, w_K\}$ where $K \in$ set of words provided by a caption corresponding to a visual sample; <br>     Target entity set $T = \{w_1, w_2, w_3, \ldots\ldots\ldots, w_N\}$ where caption with a subsequence of $N$-word represents the target entity and; |
| **Output:** Categorization of sentiment label $Y \in \{positive, negative, neutral\}$ based on target $T$ |
| 1. for $E \leftarrow 1$ to *Epochs* do <br><br>     $\mathcal{F} = \{\mathcal{F}_1, \mathcal{F}_2, \mathcal{F}_3, \ldots\ldots\ldots, \mathcal{F}_J\} \leftarrow I$ extraction of faces from the visual set using **Eq. (1)**; <br>     $D = \{D_1, D_2, D_3, \ldots\ldots\ldots, D_\mathcal{F}\} \leftarrow \mathcal{F}$ generation of fluent linguistic emotional face descriptions using extracted faces obtained from the previous step; <br>     $D_T \leftarrow D$ refined face descriptions are obtained concatenated with the target entity using **Eq. (2) to Eq. (6)**; <br>     $I_C = \{I_{C1}, I_{C2}, I_{C1}, \ldots\ldots\ldots, I_{CG}\} \leftarrow I$ generate captions from the visual modality using **Eq. (7)**; <br>     $[CLS]w_1^C, \ldots, w_K^C[SEP]w_1^T, \ldots, w_N^T[SEP]w_1^{D_T}, \ldots, w_L^{D_T}[SEP]$ fine-tune the combination of available text, target and refined facial descriptions using robustly optimised language model using **Eq. (8)**; <br>     $[CLS]w_1^C, \ldots, w_K^C[SEP]w_1^T, \ldots, w_N^T[SEP]w_1^{I_C}, \ldots, w_G^{I_C}[SEP]$ fine-tune the combination of available text, target and generated captions using robustly optimised language model using **Eq. (9)**; <br>     **Eq. (8)** ⊙ **Eq. (9)** gated fusion of the result obtained from the previous two steps; <br>     $Y \in \{positive, negative, neutral\}$ by passing **Eq. (8)** ⊙ **Eq. (9)** by a fully connected layer and softmax layer; <br>     Calculate the loss $\mathbb{L} = -\frac{1}{|D|}\sum_{\ell=0}^{|D|} \log \mathcal{P}\{Y^\ell | O^\ell\}$ and perform backpropagation; <br> 2. **end** |

For the second challenge, facial attributes obtained from the previous step are transformed into the textual representation without training an additional visual-caption model. The prediction confidence score $\alpha$ is used to pick out facial attributes. The face attributes that have a score below the threshold $\alpha = 0.5$ are filtered out. We manually develop a visual feature pattern to create fluent linguistic emotional face descriptions $D = \{D_1, D_2, D_3, \ldots\ldots\ldots, D_\mathcal{F}\}$. An example of facial description generation is shown in **Table 3**.

Table 3 A fluent face description, for example, 1 of **Figure 2**, may be generated via a template.

| Template | Instances | Visual features | | | | Fluent linguistic emotional face descriptions |
|---|---|---|---|---|---|---|
| | | Age | Gender | Race | Sentiment | |
| A [Race] [Gender] with [Age]-year-of-age exhibits [Sentiment] expression | Example 1 | 43 | man | Black | negative | A Black man with 43 years of age exhibits a negative expression |
| | Prediction confidence score | 1.00 | 1.00 | 0.879 | 0.9467 | |

### 3.2.2 Target Alignment and Refinement of the Face Descriptions

Sometimes, a multi-face image sample with varied facial expressions fails to estimate the correct emotion of the target entity. On the other hand, the inclusion of redundant facial emotions produces noise and diminishes the overall effectiveness. Therefore, it is essential to effectively synchronise the facial emotions shown in the visual sample with the desired target entity. In our proposed framework, VECTN, this component focuses on aligning facial expressions with the target object, resulting in more detailed facial descriptions.

The TDMSA challenge needs external visual-caption alignment information for more fine-grained alignments due to restricted dataset size and the absence of direct visual-caption alignment supervision. Hence, to achieve fine-grained alignment, we use caption and visual encoders of a recently developed contrastive visual-caption pre-training architecture trained on a variety of visual-caption pairs [47] denoted as '$\tau$' to encode the face descriptors $D$ associated with target $T$ and the visual $I$. The resulting embeddings for images and descriptions of faces are shown in **Equation (2)** and **Equation (3)**, where '$\oplus$' denotes concatenation.

$$O_{D_T} = Caption\_Encoder\ (\tau)(D \oplus T) \tag{2}$$
$$O_I = Visual\_Encoder(\tau)(I) \tag{3}$$

Subsequently, the obtained feature embeddings are projected into the same feature space. Then, we compute the Levenshtein distance $L$ for these feature embeddings using L2-normalization $'\gamma'$. Next, we choose and regenerate the face description that best fits the current image as the visual, emotional clue for the current target based on the similarity score using $L$. The refined face description only contains the target object and expressions based on predicted facial traits.

$$O'_{D_T} = \gamma(O_{D\&T} \cdot V_{D\&T}) \tag{4}$$
$$O'_I = \gamma(O_I \cdot V_I) \tag{5}$$
$$L = O'_I \cdot (O'_{D_T}{}^\mathbb{T}) * e^t \tag{6}$$

$V_{D\&T}$ and $V_I$ are trainable weights, and $t$ is the scaling factor of the generative visual-to-caption transformer model [44]. This module is only used for multi-face visual samples. The target is concatenated directly with the acquired face description in this module. Subsequently, the newly concatenated phrases are used as input for the textual encoder of '$\tau$', while the picture is employed as input for the visual encoder of '$\tau$'. The cosine similarities between the visual and textual features are then computed using **Equation (4), (5)** and **(6)**. Then, we choose the facial description with the highest score and modify it to get a more refined description.

Furthermore, considering the effect of image scene details on multimodal semantics, we use a recently developed, more effective generative transformer [44] for visual-to-caption translation '$\delta$' to provide an overall comprehensive description of all the image samples of the dataset using **Equation (7)**.

$$I_C = \delta(I) \tag{7}$$

Ultimately, we achieve the accurate alignment of face descriptions and visual captions, which are then used as input for the succeeding module.

### 3.2.3 Fusion Module

This module aims to combine already available caption($C$), target entity ($T$), refined facial description ($D_T$), and the generated caption ($I_C$). To leverage the pre-trained language model's robust textual context analysis, we concatenate the refined face descriptions and image caption with available text and target to create two new phrases as shown in **Equation (8)** and **(9)** below:

$$[CLS]w_1^C, \ldots, w_K^C[SEP]w_1^T, \ldots, w_N^T[SEP]w_1^{D_T}, \ldots, w_L^{D_T}[SEP] \tag{8}$$

$$[CLS]w_1^C, \ldots, w_K^C[SEP]w_1^T, \ldots, w_N^T[SEP]w_1^{I_C}, \ldots, w_G^{I_C}[SEP] \tag{9}$$

Fine-tuning two robustly optimised per-trained language models [48] with these new phrases yields $[CLS]$ token $O_{D_T}^{[CLS]} \in R^{768}$ and $O_{I_C}^{[CLS]} \in R^{768}$ pooler outputs. The gate mechanism is used to reduce noise in feature representations of $O_{D_T}^{[CLS]}$ and $O_{I_C}^{[CLS]}$. At last, to predict sentiment, fused feature representations (**Equation (10)** and **(11)**) are sent via a linear classifier using **Equation (12)**, where $V_{D_T}$, $V_{I_C}$ and $V$ are trainable weights of dimensions $R^{768\times768}$, $R^{768\times768}$, and $R^{768\times3}$. In contrast, $b_j$ and $b$ are learnable biases with dimensions $R^{768}$ and $R^3$.

$$jt = tanh\left(V_{D_T}O_{D_T}^{[CLS]} + V_{I_C}O_{I_C}^{[CLS]} + b_j\right) \tag{10}$$

$$O = jt * O_{D_T}^{[CLS]} + jt * O_{I_C}^{[CLS]} \tag{11}$$

$$\mathcal{P}(Y|O) = Softmax\big((V * O) + b\big) \tag{12}$$

All module parameters are optimised using conventional cross-entropy loss defined in **Equation (13)**.

$$\mathbb{L} = -\frac{1}{|D|}\sum_{\ell=0}^{|D|} \log \mathcal{P}\{Y^\ell|O^\ell\} \tag{13}$$

## 4 Experimental Setup and Results

### 4.1 Experimental Details

Our model was trained and evaluated using two publically accessible benchmark datasets, Twitter-2015 and Twitter-2017. Both datasets include tweets consisting of visual-caption pairs. Each caption has been tagged with some target entity and its associated sentiment polarity. Our approach primarily emphasises cases that include facial images. Therefore, we extract samples with facial images from the aforementioned two datasets to create the Tweet1517-Face dataset. Subsequently, we evaluate the effectiveness of our proposed model on these samples to prove its superiority over the others. The comprehensive statistical information for the three datasets can be seen in **Table 4**.

Table 4 Statistical information that describes all the datasets utilised in the evaluation of our proposed model.

| Name of the Dataset | Positive Samples | | | Negative Samples | | | Neutral Samples | | | Average Number of Targets | | |
|---|---|---|---|---|---|---|---|---|---|---|---|---|
| | Train | Valid | Test | Train | Valid | Test | Train | Valid | Test | Train | Valid | Test |
| **Twitter-15** | 928 | 303 | 317 | 368 | 149 | 113 | 1883 | 679 | 607 | 1.34 | 1.33 | 1.35 |
| **Twitter-17** | 1508 | 515 | 493 | 1638 | 517 | 573 | 416 | 144 | 168 | 1.41 | 1.43 | 1.45 |
| **Tweet1517-Face** | 1285 | 449 | 442 | 408 | 137 | 156 | 1531 | 514 | 494 | 1.37 | 1.37 | 1.39 |

Additionally, the model's learning rate has been configured to be 2e-5. The batch size has been set to 32, and a dropout rate of 0.4 has been employed. The proposed approach has undergone fine-tuning for a total of 15 epochs. This work was implemented using the PyTorch framework and is executed on a high-end NVIDIA TITAN RTX (48GB) GPU system with an Intel Xeon Silver 4116 CPU, 10TB storage, and 128 GB RAM. The final result has been determined by calculating the mean of five independent training iterations.

### 4.2 Baselines for Comparison

This section compares our proposed model with several cutting-edge baseline approaches (discussed in **section 2.2**) for the task of TDMSA for both unimodal and multimodal networks. **Table 5** demonstrates that the experimental results of our proposed model are more accurate than those of the other baseline methodologies in terms of Accuracy (A) and macro − F1 score, thus proving our model's superiority.

Table 5 Experimental results of our proposed model compared with the multimodal baseline approaches.

| Methods (visual + captions) | Twitter-2015 | | Twitter-2017 | |
|---|---|---|---|---|
| | A | macro − F1 | A | macro − F1 |
| TomBERT [31] | 76.18 | 71.27 | 70.50 | 68.04 |
| ESAFN [33] | 73.38 | 67.37 | 67.83 | 64.22 |
| EF-Net [34] | 73.65 | 67.90 | 67.77 | 65.32 |
| ModelNet [35] | 79.03 | 72.50 | 72.36 | 69.19 |
| R-GCN [36] | - | 75.00 | - | 87.11 |
| HIMT [38] | 78.14 | 73.68 | 71.14 | 69.16 |
| FGSN [40] | 74.61 | 65.84 | - | - |
| [41] | - | 72.97 | - | 71.76 |
| EF-CaTrBERT [43] | 77.92 | 73.90 | 72.30 | 70.20 |
| **Our Proposed (VECTN)** | **81.23** | **80.61** | **77.42** | **75.19** |

**Figure** *3* Year-wise comparison of our proposed model with already existing cutting-edge multimodal networks in terms of accuracy and macro-f1 scores**Figure 3** gives a year-by-year visual representation to compare our model with the techniques that are considered to be state-of-the-art in terms of Accuracy (A) and macro − F1 score for the Twitter-2015 and Twitter-2017 datasets, respectively.

### 4.3 Analysis of Our Experimental Results

This section presents experimental results and analysis of our proposed framework for all three datasets. **Table 6**, **Table 7** and **Table 8** highlight the best scores on each performance measure for all three datasets. Our method demonstrates superior performance when compared to all other multimodal baselines. This serves as evidence of the efficacy of the proposed VECTN framework. In the fusion module, we conducted fine-tuning using RoBERTa-base and RoBERTa-Large language models [48] and found that RoBERTa-Large yielded superior results. The model we developed indicates higher accuracy when using a more robust language model. This observation highlights the significant impact of the language model's contextual modelling capacity during the fusion step. However, this observation is not evident in the comparison of the base version model. We believe the inadequate text context modelling by the pre-trained language model in the base version is the cause of this issue.

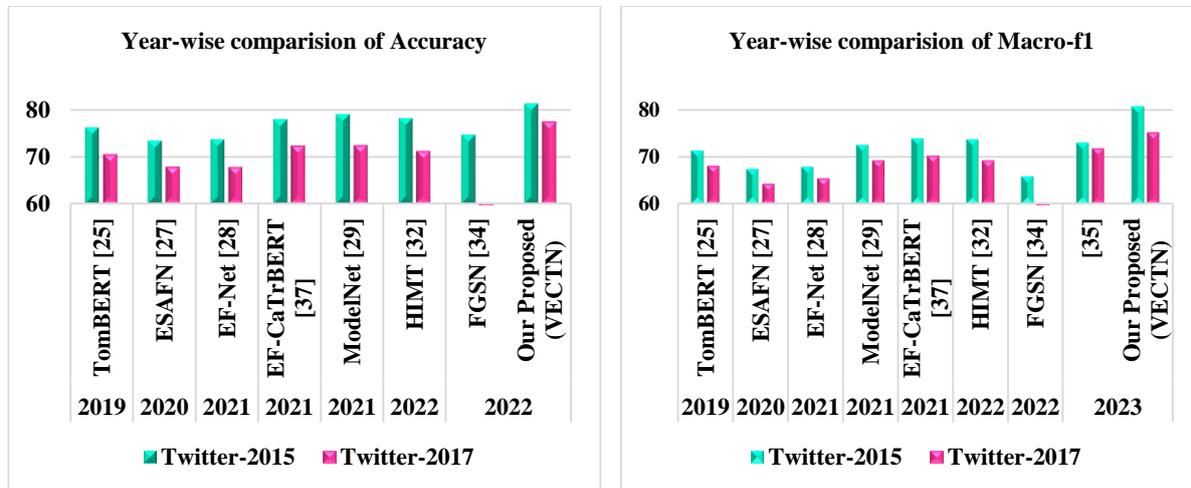

**Figure 3** Year-wise comparison of our proposed model with already existing cutting-edge multimodal networks in terms of accuracy and macro-f1 scores

**Table 6** Experimental results of our proposed approach for Twitter-2015.

| Methods | Twitter-2015 | | | | | |
|---|---|---|---|---|---|---|
| | Unimodal | | | | Multimodal | |
| | Caption | | visual | | Caption + Visual | |
| | A | macro − F1 | A | macro − F1 | A | macro − F1 |
| VECTN-Roberta-base | 73.19 | 71.74 | 75.20 | 72.51 | 79.63 | 77.21 |
| VECTN-Roberta-Large | 75.83 | 75.20 | 77.59 | 76.86 | **81.23** | **80.61** |

**Table 7** Experimental results of our proposed approach for Twitter-2017.

| Methods | Twitter-2017 | | | | | |
|---|---|---|---|---|---|---|
| | Unimodal | | | | Multimodal | |
| | Caption | | visual | | Caption + Visual | |
| | A | macro − F1 | A | macro − F1 | A | macro − F1 |
| VECTN-Roberta-base | 67.42 | 64.35 | 68.54 | 67.40 | 72.37 | 70.84 |
| VECTN-Roberta-Large | 72.11 | 70.65 | 74.48 | 71.79 | **77.42** | **75.19** |

**Figure 4** shows that combining visual and textual modes enhances performance. Therefore, this demonstrates that our proposed framework has the capability to accurately represent facial expressions shown in images. Additionally, it emphasises the need to explicitly integrate emotional cues in visual analysis.

**Table 8** Experimental results of our proposed approach for Tweet1517-Face.

| Methods | Tweet1517-Face Dataset | | | | | |
|---|---|---|---|---|---|---|
| | Unimodal | | | | Multimodal | |
| | Caption | | visual | | Caption + Visual | |
| | A | macro − F1 | A | macro − F1 | A | macro − F1 |
| VECTN-Roberta-base | 64.32 | 62.67 | 65.22 | 63.80 | 74.02 | 71.28 |
| VECTN-Roberta-Large | 70.15 | 67.46 | 74.57 | 72.30 | **79.65** | **78.16** |

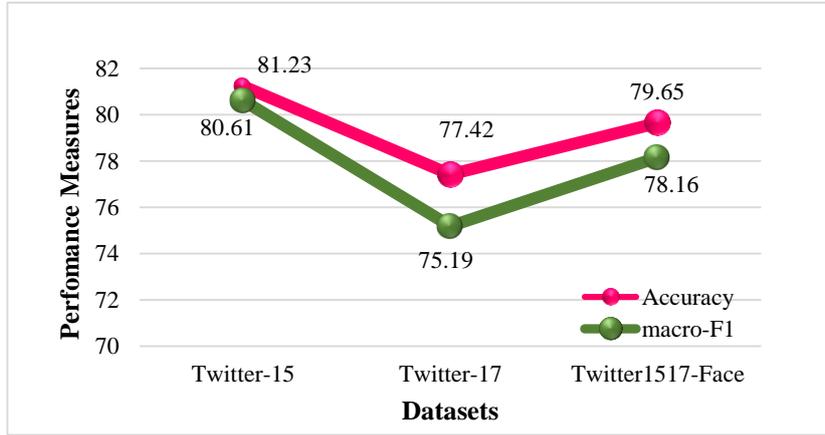

**Figure 4** Graphical representation to show the performance of the proposed model on all three datasets for multimodal (Image + Caption) configuration.

## 4.4 Ablation Study

To understand the influence of each component of our technique, we performed an extensive ablation study utilising the VECTN-RoBERTa-Large version. The results for the same are depicted in **Table 9**. Initially, the sentiment label of the target is predicted by combining the results of linguistic models. These outcomes are then used as input to a linear classification layer without the inclusion of a gating mechanism. The overall performance of the architecture experiences a significant decrease as a result of the presence of noise during the facial emotion description module. On Twitter 2015, A and macro − F1 scores dropped 1.83% and 1.96%, respectively. On the Twitter-2017 dataset, A drops 2.15%, and the macro − F1 score drops 2.88%. This suggests that the gating technique is responsible for reducing noise and extracting more useful features. Additionally, it is seen from **Table 9** that the exclusion of the target alignment module also results in a decrease in performance. This result suggests that the alignment between the visual and emotional cues and the target entity is crucial. At last, we analyse the impact of excluding the visual caption from the scene, which results in a significant decrease in the model's performance. This finding provides evidence that the utilisation of visual-to-caption translation contributes to the advancement of visual-caption fusion.

**Table 9** Ablation study conducted on Twitter-2015, Twitter-2017 and Tweet1517-Face

| Methods | Twitter-2015 | | Twitter-2017 | | Tweet1517-Face | |
|---|---|---|---|---|---|---|
| | Multimodal (Caption + visual) | | Multimodal (Caption + visual) | | Multimodal (Caption + visual) | |
| | A | macro − F1 | A | macro − F1 | A | macro − F1 |
| **VECTN-Roberta-Large** | 81.23 | 80.61 | 77.42 | 75.19 | 79.65 | 78.16 |

| Methods | Twitter-2015 Multimodal (Caption + visual) | | Twitter-2017 Multimodal (Caption + visual) | | Tweet1517-Face Multimodal (Caption + visual) | |
|---|---|---|---|---|---|---|
| | A | macro−F1 | A | macro−F1 | A | macro−F1 |
| VECTN without gating mechanism | 79.40 | 78.65 | 75.27 | 72.31 | 77.29 | 76.08 |
| VECTN without Target Alignment | 79.11 | 79.20 | 76.44 | 73.60 | 78.31 | 76.82 |
| VECTN without visual caption of the scene | 78.89 | 78.50 | 75.23 | 72.56 | 77.48 | 75.27 |

## 5 Predictive Analysis of Few Samples

In order to provide a more comprehensive demonstration to highlight the benefits of the proposed method, this section of the manuscript will include the actual predictions made by our model on a few samples gathered from Twitter-15 and Twitter-17. As shown in **Table 10**, the VECTN model accurately forecasts positive sentiments for the target terms [*LeBron James's*], [*JordanStrack*], negative sentiments for aspect words [*Harriette*], [*Anthony Kiedis*], while the neutral sense of emotion for [*Donald Trump Republcian*]. As a result, this study demonstrates that the proposed approach efficiently focuses on multimodal sentimental regions to leverage the interaction between the image and the target phrase more extensively than existing methods. Hence, the VECTN model can deeply examine the local semantic relationship between image and text in contrast with baseline models. In simple terms, the proposed model exhibits a higher degree of advantage.

**Table 10** Sentiment prediction by employing the proposed model "VECTN" on a few multimodal samples of Twitter-15 and Twitter-17.

| Modalities | | VECTN Prediction |
|---|---|---|
| Text | Image | |
| *Finals*: [*LeBron James's*]$_{Positive}$ *Record Has Improved with Age NBA* | 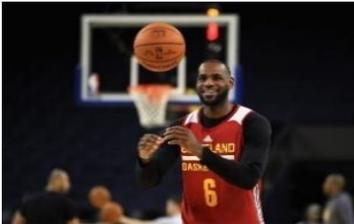 | *LeBron Jame's* − *Positive* |
| [*JordanStrack*]$_{Positive}$: *presented the che for winning the 2015 Marathon Classic. Chella Choi* | 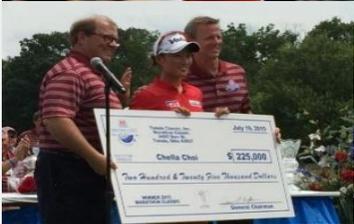 | *JordanStrack* − *Positive* |
| *Just go ahead*: *64 % of likely* [*Donald Trump Republican*]$_{Neutral}$ *voter say Paul Ryan should endorse* | 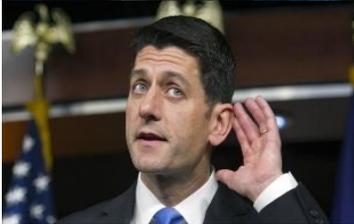 | *Donald Trump Republican* − *Neutral* |

| Modalities | | VECTN Prediction |
|---|---|---|
| Text | Image | |
| [Harriette$_{Negative}$] moved back to Chicago to care for her mom: And it's been terrible | 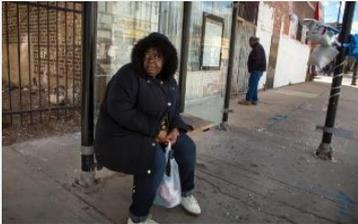 | Harriette − $Negative$ |
| [Anthony Kiedis$_{Negative}$] illness could affect Rock on the Range festival | 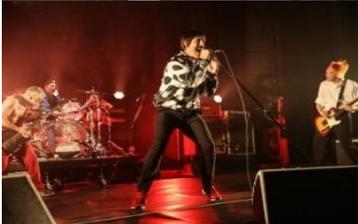 | Anthony Kiedis − $Negative$ |

## 6  Conclusion and Future Direction

This paper presents a target-dependent multimodal sentiment recognition strategy called a visual-to-emotional-caption translation network. The idea put forward utilises facial emotions depicted in images as visual indicators of emotions. In this study, we propose a novel and efficient approach to establish a correlation between the target entity in textual information and the facial expressions depicted in visual media. Our approach has successfully achieved ground-breaking results on the Twitter2015 and Twitter-2017 datasets. The results indicate that our proposed solution surpasses a set of baseline models. This showcases the strength of our method in gathering emotional clues from the visual modality and achieving cross-modal alignment on visual-caption sentimental information. In the future, we would like to extend our proposed method for other multimodal tasks, such as the identification of hate speech, sarcasm, fake news, etc. The analysis of emotions conveyed by video is another exciting field of study with promising future prospects.